\documentclass{llncs}
\usepackage{lineno,hyperref}

\usepackage{subcaption}
\usepackage{amsmath}
\usepackage{amssymb}
\usepackage[table,xcdraw]{xcolor} 
\usepackage{mathtools}

\begin{document}

\begin{frontmatter}

\title{Synaptic partner prediction from point annotations in insect brains}

\author{Julia Buhmann\textsuperscript{1,*}, Renate Krause\textsuperscript{1,2,*}, Rodrigo Ceballos Lentini\textsuperscript{1}, Nils Eckstein\textsuperscript{1}, Matthew Cook\textsuperscript{1}, Srinivas Turaga\textsuperscript{2}, Jan Funke\textsuperscript{2,3}}
\institute{\textsuperscript{1}Institute of Neuroinformatics UZH/ETHZ, Zurich, Switzerland  \linebreak
	\textsuperscript{2}HHMI Janelia Research Campus, Ashburn, USA  \linebreak 
	\textsuperscript{3}Institut de Robotica i Informatica Industrial, UPC, Barcelona, Spain  
	\linebreak
		\textsuperscript{*}corresponding authors: {\tt \{juliab,rekrau\}@ini.uzh.ch} (equal contribution) 
		\linebreak}

\maketitle

\begin{abstract}
High-throughput electron microscopy allows recording of lar\-ge stacks of neural tissue with sufficient resolution to extract the wiring diagram of the underlying neural network. Current efforts to automate this process focus mainly on the segmentation of neurons. However, in order to recover a wiring diagram, synaptic partners need to be identified as well. This is especially challenging in insect brains like \emph{Drosophila melanogaster}, where one presynaptic site is associated with multiple postsynaptic elements. Here we propose a 3D U-Net architecture to directly identify pairs of voxels that are pre- and postsynaptic to each other. To that end, we formulate the problem of synaptic partner identification as a classification problem on long-range edges between voxels to encode both the presence of a synaptic pair and its direction. This formulation allows us to directly learn from synaptic point annotations instead of more expensive voxel-based synaptic cleft or vesicle annotations. We evaluate our method on the MICCAI 2016 CREMI challenge and improve over the current state of the art, producing 3\% fewer errors than the next best method.

\end{abstract}


\end{frontmatter}


\section{Introduction}
The field of Connectomics studies the reconstruction and connectivity of neurons. So far, only electron microscopy (EM) methods are able to image neural tissue with a resolution high enough to resolve connectivity on a synaptic level. This results in large amounts of data, requiring automated methods in order to be processed in a reasonable amount of time. In order to acquire a full connectivity diagram we need to solve two main problems: First, the neurons have to be segmented to reconstruct their morphology. Second, synaptic connections between neurons have to be identified, since neuron morphology does not provide enough information to estimate connectivity \cite{kasthuri2015saturated}.

\begin{figure}[t!]	
	\centering
	\begin{subfigure}{1.0\textwidth}
		\includegraphics[width=\textwidth]{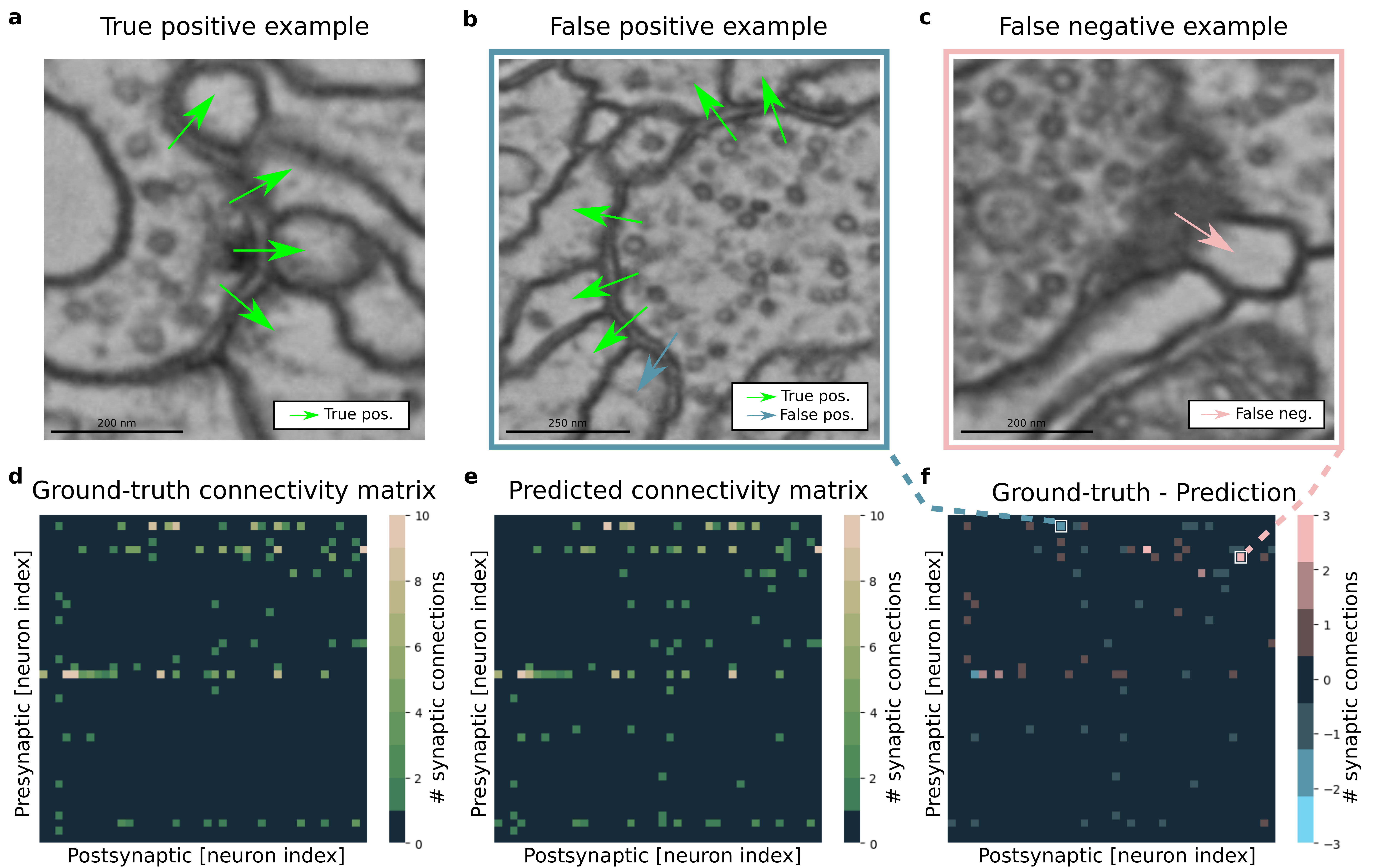}
	\end{subfigure}
	\caption{Qualitative results on the validation set. First row: examples of \textbf{(a)} True Positive (TP), \textbf{(b)} False Positive (FP) and \textbf{(c)} False Negative (FN) in 2D sections. Note that in (b) the predicted partner is likely correct and was overlooked during ground truth annotation while (c) represents an ambiguous case; Second row: Connectivity matrices of \textbf{(d)} ground truth and \textbf{(e)} predicted connectome. Each row and each column in a connectivity matrix corresponds to one neuron. A positive entry in the matrix stands for the number of connections/synapses there exists between the two neurons. In \textbf{(f)} the ground truth matrix is subtracted from the prediction matrix (blue: FP, red: FN). Results shown for the validation set (f-score 0.75) and ground truth segmentation based synaptic partner extraction.}
	\label{fig:connectivitymatrix}
\end{figure}

The segmentation of neurons has received considerable attention by the computer vision community, for recent methods see \cite{funke2017large,januszewski2016flood}. Automated synapse detection methods also have seen progress, for instance in   \cite{dorkenwald2017automated}, automatically identified synapses allow the analysis of different neuron cell types. Most of the proposed methods identify synaptic \emph{sites} \cite{becker2013learning,jagadeesh2014synapse,kreshuk2014automated,kreshuk2011automated,roncal2014vesicle}. For the overall goal of reconstructing the connectome, however, it is necessary to additionally identify the pre- and postsynaptic \emph{partners}. In mouse, zebrafish or zebra finch a synaptic site usually has only one pre- and postsynaptic partner, whereas in insects such as \textit{Drosophila melanogaster}, synapses form one-to-many or connections which are called polyadic synapses (for an example, see Fig. \ref{fig:connectivitymatrix} a and b).

To identify partners in one-to-one synapses, a common strategy is to first identify contact sites of neuronal segments to extract a set of potential synaptic partners, and second to collect features for them which are then handed over to a classifier to decide if a given candidate is valid \cite{dorkenwald2017automated,staffler2017synem}. 

In \cite{huang2016fully}, a similar strategy is used to detect synaptic partners in polyadic synapses found in insects. Specific changes are required to make the method applicable. For instance, while a vesicle cloud is quite indicative for a presynaptic site in song bird \cite{dorkenwald2017automated}, the \textit{Drosophila melanogaster} presynaptic site includes a characteristic T-shaped appearance (called T-bar). In a first step, a convolutional neural network is trained to detect T-bars. Subsequent partner prediction is constrained to those identified locations. This method was shown to perform well on isotropic data. 

However, for anisotropic datasets, the task of automatically detecting presynaptic locations is potentially harder. 
The only currently proposed method to detect synaptic partners in anisotropic datasets with polyadic synapses by \cite{kreshuk2015talking} requires the neuron segmentation as well as predicted synapse locations. Based on these, the method identifies a set of potential synaptic partners. Then unary and pairwise factors are assigned to all candidate pairs and an Integer Linear Program (ILP) is solved to select correct synaptic partners.

\section{Method} \label{methods}
A common approach for synaptic partner detection is to first identify potential pre- and postsynaptic point locations while the correct pairing of candidate sites is performed in a second, separate step \cite{dorkenwald2017automated,kreshuk2015talking,staffler2017synem}. We here propose a method for synaptic partner prediction that fuses these two, formerly independent steps, into one. We achieve this reduction by formulating the problem as a classification task on the space of directed edges between voxels. An advantage of this approach is that potential pre- or postsynaptic sites are only classified as such if there is evidence for a corresponding partner. The proposed representation also allows us to learn from synaptic point annotations only, since we do not rely on labeled synaptic features, such as synaptic clefts or vesicle clouds.   

In the following subsections, we describe how we 1) represent neuron connectivity directed long-range edges, 2) train a U-Net to classify those edges, and 3) post-process the U-Net output to directly obtain synaptic partners.

\subsection{Directed edges for synaptic partner representation} \label{subsec:representation}
In order to constrain the learning problem we fuse synaptic site detection and partner identification in a single step. To this end we consider a directed graph $G(V,E)$, where each voxel is represented as a vertex $v\in V$ and we encode relations between voxels as the directed edges $e_{i \rightarrow j} = (v_{i}, v_{j}) \in E$.

If we were to connect each voxel $v_{i}$ in a given volume with each other voxel $v_{j\neq i}$, we could identify synaptic partners by selecting the subset of those edges $e_{i\rightarrow j} = (v_{i}, v_{j})$ whose vertices $v_{i}$ and $v_{j}$ lie inside a pre- and corresponding postsynaptic region respectively, pointing from one site to the other. However, in such a scheme the number of candidate edges we would need to consider scales as $|V|^2$, which is unfeasible and not necessary.

We thus limit the space of candidate edges to lie within a biologically plausible range using the prior that pre- and postsynaptic sites are separated from each other by a certain distance. That is for each voxel $v_i$ we only consider $n_e$ edges with associated relative offsets $\boldsymbol{r} \in \mathbb{Z}^3$. The number of edges $n_e$ trades off the coverage rate of synaptic partners. For example, if we use too few edges, there might be synaptic partners that are missed by our representation. In order to select a minimal set of sufficient edges, we perform a grid search on a training and validation set over the number of candidate edges $n_e$, the length of the vector $|\boldsymbol{r}|$ and radius of pre- and postsynaptic region $r_\text{syn}$ such that we obtain a coverage rate of $100\%$ of synaptic partners. We denote the set of edges, that we find with the proposed method $E^\text{syn}$.

\begin{figure}[t!]
	\centering
	\includegraphics[width=0.95\textwidth]{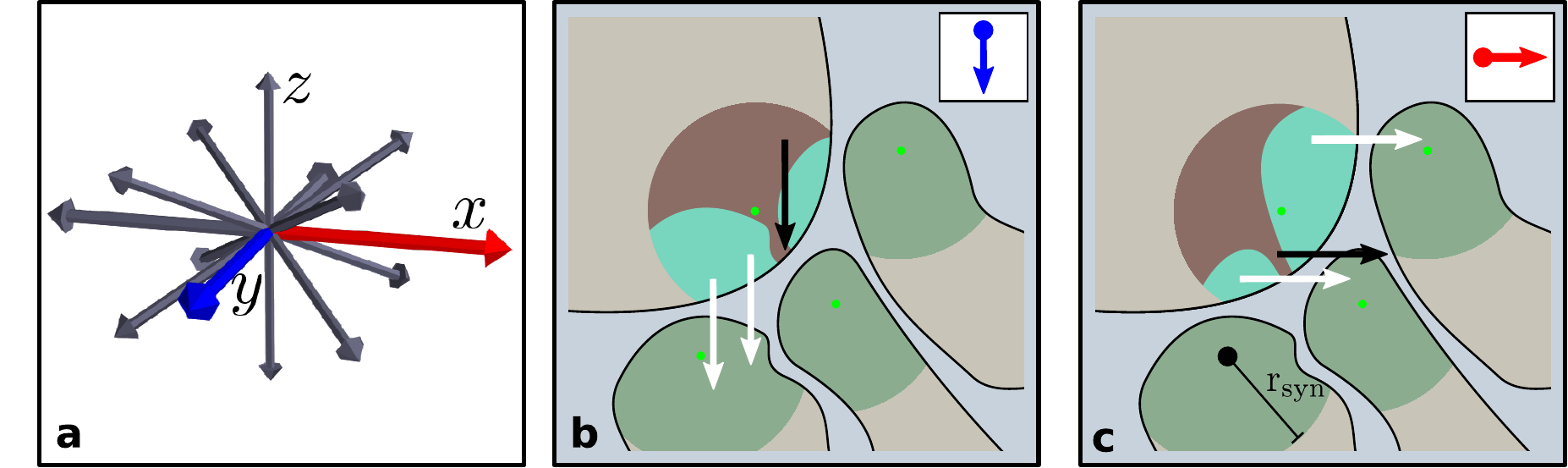}
	\caption{Illustration of synaptic partner representation. \textbf{(a)} 3D rendering of the 14 offset vectors $\boldsymbol{r}$ found with grid search on the CREMI dataset; \textbf{(b)} and \textbf{(c)} One presynaptic neuron with its 3 postsynaptic partners. An example is shown for two offset vectors (b) $\boldsymbol{r}=(0, 120, 0)$, (c) $\boldsymbol{r} = (120, 0, 0) $ in the x-y plane, their direction shown in upper right corner and marked in (a) in blue and red. Green dots represent synaptic point annotation, with their pre- (brown) and post- (green) synaptic location area. White arrow: selected edge $s(e_{i \rightarrow j}=1)$, black arrow: unselected edge $s(e_{i \rightarrow j}=0)$. Teal colored regions correspond to the resulting activation map: each voxel $v_{i}$ is colored teal if the corresponding edge $e_{i \rightarrow j}$ is selected, i.e. voxel $v_{j}$ lies inside the postsynaptic area.}
	\label{fig:affinity_examples}
\end{figure}

\subsection{Edge classification}
We train a 3D U-Net to classify for each voxel $v_{i}$ which of its directed edges $e_{i \rightarrow j} \in E^{\text{syn}}$ is synaptic or non-synaptic. That is we predict a $n_e$ dimensional vector $\vec{y}^{\text{i}}$ for each voxel in the input volume. Each entry $\vec{y}^{i}_{j}$ then represents the score $s(e_{i \rightarrow j} = 1)$ that edge $e_{i \rightarrow j}$ is synaptic. Consistent with the derivation of edges $e \in E^\text{syn}$ we annotate synaptic regions for training by expanding the ground truth point annotation of pre- and postsynaptic sites into a ball of radius $r_{\text{syn}}$, shown in brown/green in Fig. \ref{fig:affinity_examples}. Note that we constrain each synaptic regions to lie within its corresponding neuron.  An illustration of pre- and postsynaptic regions, selected and unselected edges for two offset vectors $\boldsymbol{r}$ in the x-y plane is shown in Fig. \ref{fig:affinity_examples}.     

\subsection{Synaptic partner extraction}\label{sec:pextraction}
We formalize the notion of a connectome by introducing a directed connectome graph $G_{c}=(S,A)$, with vertices $s \in S \subset P(V)$ representing neuron segments as subsets of voxels $v \in V$ and candidate synapses between neurons $a \in A \subset P(E)$ as subsets of edges $e \in E$, where $P(\cdot)$ is the power set. We extract candidate synapses $a_{k \rightarrow l}$ from the edge predictions in the following way: we use an underlying neuron segmentation (this can be either ground truth or automatically generated) to collect all edges $e$ that start in segment $s_{k}$ and end in another segment $s_{l \neq k}$. We only consider edges as evidence for a synapse $a$ if the predicted edge score $s(e=1)$ is above a certain threshold $t_{1}$. In order to account for neuron pairs that have multiple synapses at different locations, we introduce a separate candidate synapse $a^{m}_{l \rightarrow k} \subset a_{l \rightarrow_k}$ for each set of edges whose target voxels form a connected component in segment $s_{k}$. A confidence score for candidate synapses $a^{m}_{l \rightarrow k}$ is obtained by calculating the sum over all edge scores per synapse:
\begin{equation}\label{eq:constrainteq1}
\text{confidence}({a^{m}_{l \rightarrow k}}) = \sum_{e\in\boldsymbol{a^{m}_{l \rightarrow k}}}s(e)
\end{equation}
A candidate synapse $a^{m}_{l \rightarrow k}$ is finally selected if confidence($a^{m}_{l \rightarrow k}) > t_{2}$. The parameters $t_{1}$ and $t_{2}$ are used to control precision and recall of the method. Finally we extract single locations for the identified partners by calculating the center of mass of the start and end points of synapses $a^m_{l \rightarrow k}$. Synaptic partners are thus represented by a single pre- and a single postsynaptic location.

\paragraph{\textbf{Error metric}}\label{sec:evaluation}
We evaluate our method using the synaptic partner identification metric proposed by the MICCAI 2016 CREMI challenge\footnote{\label{cremicit}https://cremi.org}. First it uses Hungarian Matching to match each predicted synaptic pair $a^{m}_{l \rightarrow k}$ to a ground-truth pair within a tolerance distance $d$. It further requires that the underlying segment IDs of ground-truth and predicted pair match. This means that a slight shift of the location is only tolerated if the underlying voxel still corresponds to the correct segment.

\section{Results}

\paragraph{\textbf{CREMI Dataset}}
We tested our method on the publicly available, serial section, EM dataset from the \mbox{MICCAI} 2016 \mbox{CREMI} challenge\textsuperscript{\ref{cremicit}}. The dataset is from \textit{Drosophila melanogaster} brain tissue comprising 3 training volumes A, B and C together with 3 test volumes A+, B+ and C+, with a size of 1250x1250x125 voxels each and a highly anisotropic resolution of 4x4x40~nm. We split the three training volumes into a training set (A, B and half C) and a validation set (the other half of C). The validation set is used for model optimization and qualitatively analyzing results. The ground truth for the training set includes volumetric neuron labels and synaptic partner annotations. Each partner is represented with a pre- and postsynaptic location (as shown in Fig. \ref{fig:connectivitymatrix} a). We use the training volumes to carry out a grid search in order to obtain parameters for our synaptic representation (discussed in \ref{subsec:representation}). We find $r_\text{syn}=100$~nm and 14 edges $e_{i\rightarrow j}$ per voxel with following vectors: $\boldsymbol{r} = (0, 0, \pm80)$, $(\pm120, 0, 0)$, $(0, \pm120, 0)$, $(\pm40, \pm60, \pm40)$ (offset vectors displayed in Fig. \ref{fig:affinity_examples} a).

\paragraph{\textbf{Architecture}}\label{sec:unet}
Our architecture is based on the U-Net architecture proposed by \cite{ronneberger2015u}. It has been shown to be well suited to a number of applications in biomedical imaging and neuron segmentation in particular (e.g. \cite{funke2017large}). We use the same U-Net architecture as in \cite{funke2017large} and we do not carry out any further experiments with regards to the hyperparameters of the network. We use three down- and three up-sampling layers where each layer consists of two convolutions with kernel size (3x3x3). We downsample via max-pooling operations of size (3x3x1) and stride 3 and upsample via transposed convolutions of the same size. Lastly we apply a convolutional layer with kernel size 1 and sigmoid activation function. Here, each voxel has a context of x=212, y=212, z=28 voxel (848, 848, 1120~nm). We use the Adam optimizer \cite{DBLP:journals/corr/KingmaB14} with a learning rate of $\alpha = 5 \cdot 10^{-6}$ for all our experiments.

\paragraph{\textbf{CREMI: Challenge Results}}
We include in our training the whole sample A, B, and C in order to make use of all available training data. We carry out a grid search on sample A, B, C and optimize for f-score to obtain $t_1=0.5$ and $t_2=2500$. For the synaptic partner extraction we use an automatically generated neuron segmentation (provided by the authors of \cite{funke2017large}). Results for the CREMI challenge are summarized in Table~\ref{table:cremichallenge}. We obtain a 3\% gain in the f-score over the previous challenge leader. Looking at the total number of FN and FP this translates to a clear improvement in accuracy of $\sim$ 19\%. Note that the f-score is averaged over all three samples equally, although sample A+, B+ contain presumably fewer synapses (the actual TP numbers are not revealed in the challenge). This explains the discrepancy of the $\sim$ 3\% gain in accuracy for the f-score and the 19\% gain in accuracy for the absolute numbers.

\begin{table}[t!]
	\centering
	\caption{Results on the CREMI challenge for our method (lr U-Net) compared to the next best method (PSP unar). Highlighted in bold are our method and best values per column. Reported measures are 1) f-score, 2) FP and FN averaged over the three test datasets and 3) FN and FP summed over the three test datasets. For a detailed description of the used error metric, see Section \ref{sec:evaluation}}.
	
	
	\begin{tabular}{llll}
		\rowcolor[HTML]{FFFFFF} 
		Method \ \          
		& f-score  \ \ \    & avg. FP/FN \ \ & abs. FP/FN       \\ \hline
		\rowcolor[HTML]{EFEFEF} 
		\begin{tabular}[c]{@{}l@{}}
			PSP unar \end{tabular} & 0.539 &  266/\textbf{281}        & 800/\textbf{843} \\
		\rowcolor[HTML]{FFFFFF} 
		\textbf{lr U-Net}  & \textbf{0.553}          & \textbf{175}/314 & \textbf{526}/942          \\ \hline
		\rowcolor[HTML]{EFEFEF}
		\footnotesize
		\small lr U-Net training & 0.669         & 223/185 & 668/554 \\     
	\end{tabular}
	\label{table:cremichallenge}
\end{table}

\paragraph*{\textbf{Qualitative Results}}
Since the ground truth is not public we cannot qualitatively analyze our results on the test set. Instead we summarize our observations for the validation set (f-score = 0.75). We find some errors in the ground truth annotations, where some synapses have not been annotated (FP) or at least are of ambiguous nature (see Fig. \ref{fig:connectivitymatrix}, b and c). Furthermore we observe that our method has problems with very small postsynaptic neurons or synapses that lie in the cutting plane (not shown). Fig. \ref{fig:connectivitymatrix} (d-f) show a visualization of our results in form of a connectivity matrix. Note that the overall appearance of the predicted matrix and the ground truth is surprisingly similar despite the comparably low precision and recall (0.78, 0.73).
\section{Discussion}
The main advantage of our method is the ability to directly predict synaptic partners. Using a state-of-the-art deep learning architecture, we currently lead the CREMI challenge. Nevertheless, the comparably low overall performance (f-score: $0.55$) demonstrates how challenging the task remains. Although we can confidently attribute some ``errors'' to incorrect labels or ambiguous cases, there are many observed cases that are truly incorrect. 

Our current synapse representation has the shortcoming of using a discretized approach: sparse, long range edges. Although we guarantee that all synapses in training and validation set are captured, synapse statistics of other parts of the dataset potentially vary. Our method might thus benefit from a continuous formulation, in which we predict x, y, z values for the direction of the postsynaptic partner. We thus propose to rephrase the classification task as a regression task in future work.

\section*{Acknowledgements}
This work was funded by the SNF grants P2EZP2\_165241 and 205321L\_160133. 
\clearpage 


\bibliographystyle{splncs03}
\bibliography{mybibfile}

\end{document}